\documentclass[journal]{IEEEtran} 

\makeatletter
\def\thickhline{%
	\noalign{\ifnum0=`}\fi\hrule \@height \thickarrayrulewidth \futurelet
	\reserved@a\@xthickhline}
\def\@xthickhline{\ifx\reserved@a\thickhline
	\vskip\doublerulesep
	\vskip-\thickarrayrulewidth
	\fi
	\ifnum0=`{\fi}}
\makeatother

\newlength{\thickarrayrulewidth}
\usepackage{epigraph}
\usepackage{longtable}

\usepackage{textcomp}

\usepackage{amssymb}

\usepackage[hidelinks]{hyperref}
\hypersetup{
	colorlinks=true,
	linkcolor=black,
	filecolor=black,
	citecolor=black,
	urlcolor=blue,
}

\usepackage{booktabs}

\usepackage{multirow}

\usepackage{color}
\usepackage[table, dvipsnames]{xcolor}
\definecolor{PyOrange}{RGB}{255, 201, 14}
\definecolor{PyBlue}{RGB}{112, 146, 190}

\definecolor{WordGreen}{RGB}{100, 136, 40}
\definecolor{WordDarkGrey}{RGB}{82, 82, 82}
\definecolor{WordRed}{RGB}{192, 80, 77}
\definecolor{WordBlue}{RGB}{0, 122, 192}

\definecolor{WordLightBlue}{RGB}{218, 238, 243}
\definecolor{WordLightGreen}{RGB}{234, 241, 221}

\definecolor{WordFillGreen}{RGB}{194, 214, 155}
\definecolor{WordFillRed}{RGB}{252, 214, 182}
\definecolor{WordFillGray}{RGB}{217, 217, 217}

\UseRawInputEncoding
\usepackage{times}
\usepackage{epsfig}
\usepackage{amsmath}
\usepackage{amssymb}
\usepackage[ruled,vlined,linesnumbered]{algorithm2e}
\usepackage{booktabs}
\usepackage{pgfplots}
\pgfplotsset{compat=1.16}
\usepackage{comment}
\usepackage{array, makecell} 
\usepackage{xcolor}
\usepackage{lettrine}
\usepackage{url} 
\usepackage{lscape}
\usepackage{tabularx}
\usepackage{colortbl}
\usepackage{hhline}
\usepackage{vcell}
\usepackage{longtable}
\usepackage{vcell}
\usepackage{rotating}
\usepackage{pdflscape}
\usepackage{sidecap}
\usepackage{neuralnetwork}
\usepackage[export]{adjustbox} 

\usepackage{acronym}
\acrodef{FCN}[FCN]{Fully Convolutional Network}
\acrodef{GAME}[GAME]{Grid Average Mean Absolute Error}
\acrodef{DL}[DL]{Deep Learning}
\acrodef{DNN}[DNN]{Deep Neural Network}
\acrodef{ML}[ML]{Machine Learning}
\acrodef{CV}[CV]{Computer Vision}
\acrodef{AI}[AI]{Artificial Intelligence}
\acrodef{CNN}[CNN]{Convolutional Neural Network}
\acrodef{RNN}[RNN]{Recurrent Neural Network}
\acrodef{GAN}[GAN]{Generative Adversarial Network}
\acrodef{JCU}[JCU]{James Cook University}
\acrodef{MAE}[MAE]{Mean Average Error}
\acrodef{MAP}[mAP]{Mean Average Precision}
\acrodef{CA}[CA]{Classification Accuracy}
\acrodef{LCFCN}[LCFCN]{Localization-based Counting loss Fully Convolutional Network}
\acrodef{IoT}[IoT]{Internet of Things}
\acrodef{MLP}[MLP]{Multi-Layer Perceptrons}

\usepackage[most]{tcolorbox}
\newtcolorbox[auto counter]{pabox}[2][]{%
colback=blue!5!white,colframe=blue!75!black,fonttitle=\bfseries,
title=Box~\thetcbcounter: #2,#1}

\newcommand{\MyPaperTitle}{A lightweight Transformer-based model for fish landmark detection}

\usepackage{tikz}

\ifCLASSOPTIONcompsoc
  \usepackage[caption=false,font=normalsize,labelfont=sf,textfont=sf]{subfig}
\else
  \usepackage[caption=false,font=footnotesize]{subfig}
\fi

\usepackage{cite}

\usepackage{graphicx}
\graphicspath{{./Graphics/}}
\DeclareGraphicsExtensions{.pdf,.jpeg,.png,.jpg}

\usepackage{amsmath}

\usepackage{array}

\usepackage{url}

\begin{document}
\title{\MyPaperTitle}

\author{
    \IEEEauthorblockN{Alzayat Saleh\IEEEauthorrefmark{1}, 
    David Jones\IEEEauthorrefmark{1}\IEEEauthorrefmark{2}, 
    Dean Jerry\IEEEauthorrefmark{1}\IEEEauthorrefmark{2}, 
    and Mostafa~Rahimi~Azghadi\IEEEauthorrefmark{1}\IEEEauthorrefmark{2}}
    
    \IEEEauthorblockA{\IEEEauthorrefmark{1}College of Science and Engineering, James Cook University, Townsville, QLD, Australia}
    
    \IEEEauthorblockA{\IEEEauthorrefmark{2}ARC Research Hub for Supercharging Tropical Aquaculture through Genetic Solutions, James Cook University, Townsville, QLD, Australia}
}

\maketitle

\begin{abstract}
Transformer-based models, such as the Vision Transformer (ViT), can outperform \acfp{CNN} in some vision tasks when there is sufficient training data.
However, \acp{CNN} have a strong and useful inductive bias for vision tasks (i.e. translation equivariance and locality).
In this work, we developed a novel model architecture that we call a Mobile fish landmark detection network (MFLD-net).
We have made this model using convolution operations based on ViT (i.e. Patch embeddings,  Multi-Layer Perceptrons). MFLD-net can achieve competitive or better results in low data regimes while being lightweight and therefore suitable for embedded and mobile devices. 
Furthermore, we show that MFLD-net can achieve keypoint (landmark) estimation accuracies on-par or even better than some of the state-of-the-art \acp{CNN} on a fish image dataset.
Additionally, unlike ViT, MFLD-net does not need a pre-trained model and can generalise well when trained on a small dataset.
We provide quantitative and qualitative results that demonstrate the model's generalisation capabilities. This work will provide a foundation for future efforts in developing mobile, but efficient fish monitoring systems and devices. 
\end{abstract}

\ifCLASSOPTIONpeerreview
\else
	\begin{IEEEkeywords}
Transformer, Computer Vision,  Convolutional Neural Networks 
Image and Video Processing,  Machine Learning, Deep Learning.
	\end{IEEEkeywords}
\fi


\section{Introduction}\label{secintro}
 
\acfp{CNN} have dominated the design of \acf{DL} systems used for computer vision tasks for many years.
However, architectures based on Transformer models, such as Vision Transformer (ViT)\cite{Dosovitskiy2020}, have been shown to outperform standard convolutional networks in many of these tasks, especially when large training datasets are available.

Inspired by the strong performance of Vision Transformers, we have investigated utilizing some of ViT's architectures using convolution operations. 
Specifically, we have studied the use of Patch Embeddings\cite{Dosovitskiy2020}, Multi-Layer Perceptrons (MLP-Mixer)\cite{Tolstikhin2021},  and Isometric architectures \cite{Sandler2019}.
In order to apply a transformer to greater image sizes, patch embeddings aggregate together small areas of the image into single input features.
Then, MLP-Mixer works directly with the patches as input, separating the mixing of spatial and channel dimensions, while keeping the network's size and resolution constant (i.e Isometric). 
In this work, we utilize these techniques to modify a standard \ac{CNN}'s architecture to a simple model that is similar in spirit to the ViT using convolutions operations, but does not need a pre-trained model and can generalise well when trained on a small dataset.

Our proposed network, which we named MFLD-net is implemented to estimate landmarks (keypoints) on the fish body to better understand and estimate its morphology.
In aquaculture, determining the morphology of fish is a frequent and essential task. It is crucial for selecting fish for culture as well as developing and testing novel fish strains. Fish hatchery employees, breeders, and geneticists all utilise morphological factors. Morphology helps identify when fish are mature enough to produce eggs or sperm. 
When determining the growth and maturity of fish, a number of specific morphological traits may be evaluated including the distance from the tip of the mouth to the posterior midpoint of the caudal fin, or the depth of the body from the  posterior base of the dorsal fin to anterior of the anal fin.
There are also several external, internal and functional aspects that should be considered. Due to the frequency and scale of fish morphology assessments, an automated method using fish images will significantly facilitate fish processing and improves aquaculture practice. 

To that end, we have developed the MFLD-net model that assists ecologists and fisheries managers in estimating the morphology and consequently predicting the size and other morphological aspects of the fish accurately and non-invasively. This provides them with the capacity to make informed management decisions. 
To evaluate our model, we use 
an image dataset of Barramundi (\textit{Lates calcarifer}), also known as Asian seabass.
We also compare our results to several baseline models to show the performance of MFLD-net. 

In summary, the contributions of this work are as follows: 
\begin{enumerate}
    \item We propose a simple \ac{CNN} network that estimates the position of known keypoints in a fixed-size fish image. 
    \item Due to our architectural innovations, our proposed model is very fast, compact and simple in structure making it suitable for resource-constrained mobile and embedded devices.   
    \item We compare our results with several baselines, including U-net\cite{Ronneberger2015}, ResNet-18\cite{He2015ResNet}, ShuffleNet-v2\cite{Zhang2018b}, MobileNet-v2\cite{MobileNetV2}, and SqueezeNet\cite{Iandola2016}.
    
    \item We provide an evaluation of our model on $60\%$ of our fish image dataset to quantify the generalisation and robustness of our simple MFLD-net model.
\end{enumerate}

\begin{figure*}[htbp]
\centering
\includegraphics[width=0.98\textwidth]{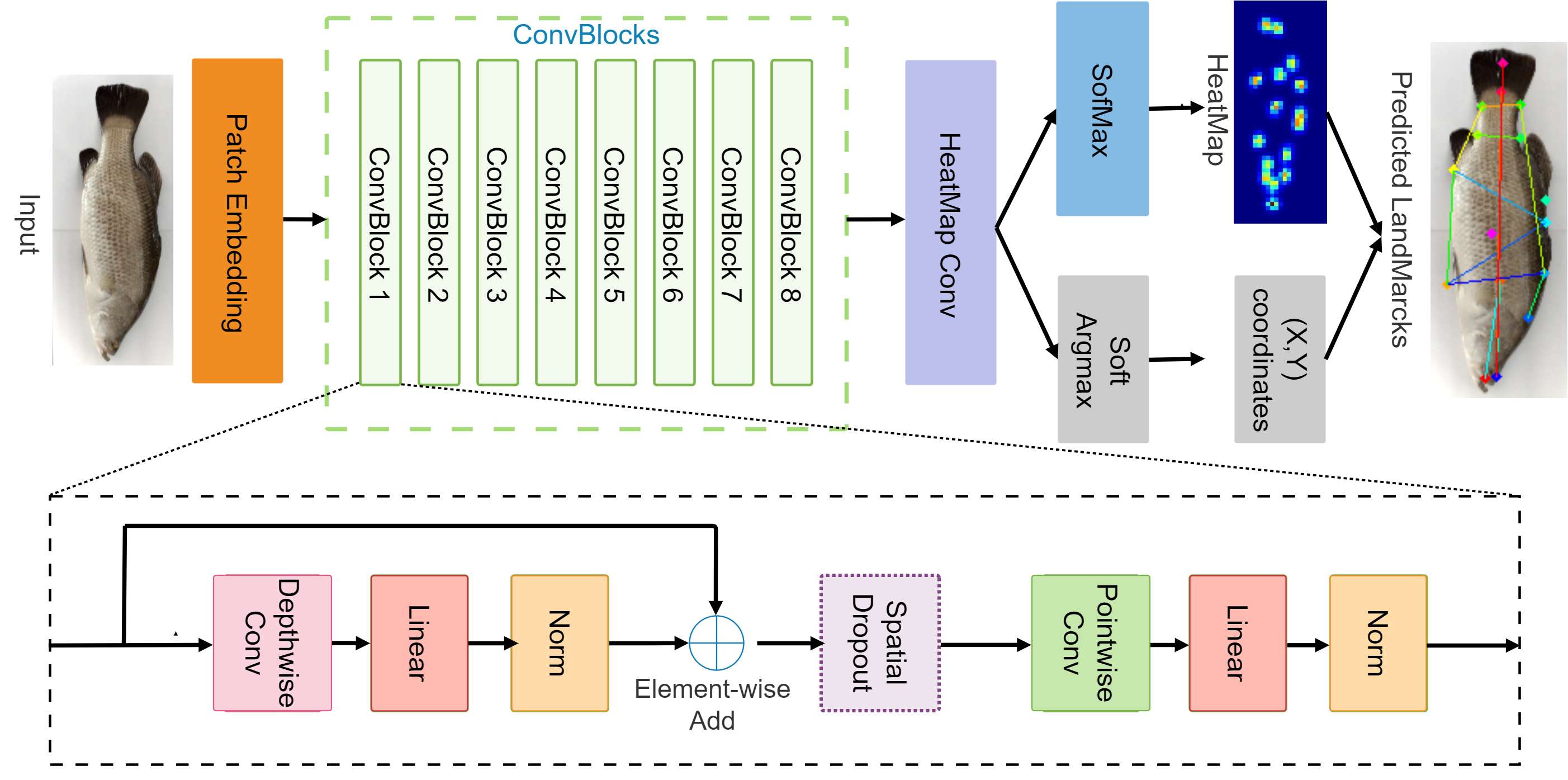}
\caption{Proposed MFLD-net architecture, which is similar in spirit to the ViTs, but uses convolutions operations for keypoints estimation.}
\label{fig:6}
\end{figure*}

\section{Related Work} 

Vaswani \textit{et al.} first \cite{Vaswani2017} suggested transformers for machine translation, and they have subsequently become the standard solution for many Natural Language Processing (NLP) applications.
Since then, there have been several attempts to incorporate convolutional network characteristics into transformers, making the Vision Transformers (ViT).
Recently, Many \acf{CV} tasks have shown that designs based on such Vision Transformer architecture\cite{Dosovitskiy2020}, outperform standard convolutional networks, particularly for big datasets.

Recently, the use of patch embeddings for the first layer has spawned a new paradigm of "isotropic" designs, i.e., those having identical sizes and shapes across the network. These models resemble repeating transformer's encoder blocks, but instead of self-attention and \acf{MLP} operations, alternative operations are used.
For example, Bello \textit{et al.}\cite{Bello2019} introduced a two-dimensional
relative self-attention mechanism replacing convolutions as a stand-alone computational primitive for image classification.
ResMLP \cite{Touvron2021} built upon multi-layer perceptrons for image classification by a simple residual network that alternates between a linear layer and a two-layer feed-forward network.

Because of its capacity to capture long-distance interactions, self-attention has been widely adopted as a computational module for modelling sequences\cite{Bahdanau2015}.
For example, Ramachandran \textit{et al.} \cite{Ramachandran2019} replaced all instances of spatial convolutions with a form of self-attention applied to a \ac{CNN} model to produce a fully self-attentional model that outperforms the baseline on ImageNet classification.

The recent popularity of pure Transformer models\cite{Dosovitskiy2020} has inspired our work to improve the overall performance of a \ac{CNN} with reduced model parameters to make it suitable for mobile applications.
Our key idea is motivated by the fact that the original Transformer uses patch embeddings and multi-layer perceptrons. We have used this idea to develop a multi-layer CNN module that has the same parameter settings as the encoder of the Transformer.

\section{Model Architecture} 

We propose the Mobile fish landmark detection network (MFLD-net), a novel end-to-end keypoint estimation model designed as a lightweight architecture for mobile devices. We apply the architecture to address some of the main issues of current methods such as accuracy and efficiency on mobile and static keypoint estimation.
The detailed architecture of MFLD-net is shown in figure \ref{fig:6}. It builds upon \acfp{CNN}\cite{Sandler2019}, Vision Transformer architecture\cite{Dosovitskiy2020}, and Multi-Layer Perceptrons (MLP-Mixer)\cite{Tolstikhin2021}.
Additionally, MFLD-net adapts a hybrid method for processing confidence maps and coordinates that provides  accurate detection for estimating keypoint locations.

To achieve higher robustness and efficiency, our architecture leverages the use of patch embedding\cite{Dosovitskiy2020}, spatial/channel locations mixing\cite{Tolstikhin2021}, as well as a combination of \acp{CNN} that have the same size and shape throughout the network, i.e. are Isometric\cite{Sandler2019}.


\subsection{Isometric  Architecture} 
Our model architecture is based on Isometric Convolutional Networks\cite{Sandler2019}, which are made up of several similar blocks with the same resolution across the model.
Architectures that are "Isometric" have the same size and shape throughout the network and maintain a fixed internal resolution throughout their entire depth (see figure \ref{fig:6}). 

Sandler \textit{et al.}\cite{Sandler2019} have demonstrated that the resolution of the input picture has only a minimal impact on the prediction quality of modern \acp{CNN}. Instead, the trade-off between accuracy and the number of multiply-adds required by the model is mostly determined by the internal resolution of intermediate tensors. Also, model accuracy can be improved further without the use of additional parameters given a fixed input resolution.

Therefore, our model has two main attributes: 
(1) No pooling layers while still maintaining a high receptive field.
(2) Isometric networks have a high degree of accuracy while needing relatively little inference memory.
These attributes make our model lightweight, hence suitable for edge processing on mobile and
low power devices, such as drones and robots, which are commonplace across various industries ranging from agriculture \cite{Lammie2019} to marine sciences \cite{Jahanbakht2022a}. 
This lightweight design does not, however, compromise accuracy due to its use of an isometric architecture. 
In an era of mobile processing \cite{Jahanbakht2021}, there is a significant need for lightweight, yet powerful and effective keypoint estimation models. 


\begin{figure}[!t]
\centering
\includegraphics[width=0.48\textwidth]{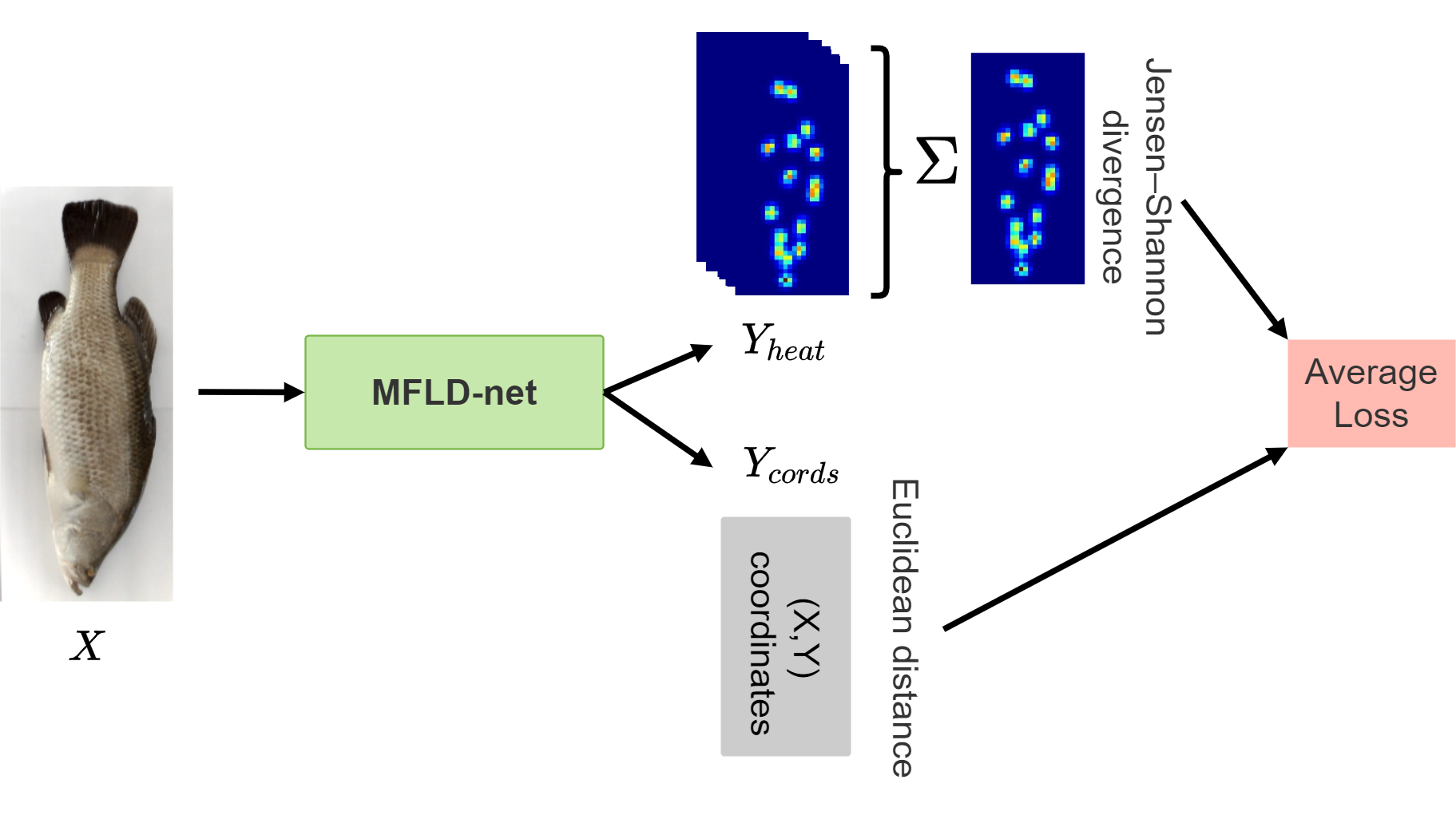}
\caption{A schematic diagram of the multi-task loss function used for training MFLD-net.}
\label{fig:10}
\end{figure}

\subsection{Patch Embedding} 
Inspired by the  Vision Transformer architecture\cite{Dosovitskiy2020}, 
we experiment with applying patch embeddings directly to a standard \ac{CNN}. 
To do so, we divide an image into patches and feed a \ac{CNN} tensor layout patch embeddings to preserve locality.
In an NLP application, image patches are processed similarly to tokens (words). Patch embeddings enable all downsampling to occur simultaneously, lowering the internal resolution and therefore increasing the effective receptive field size, making it simpler to combine sparse spatial information.
The key advantage of using \ac{CNN} instead of Transformer is the inductive bias of convolution\cite{Li2018, Cohen2017} such as translation equivariance and locality. 
Therefore, \ac{CNN}  is well-suited to vision tasks because it generalises well
when trained on a small dataset.
We implemented Patch embeddings as convolution with 3 input channels, $256$ output channels, kernel size of 4, and stride of 4, followed by 8 ConvBlocks, as can be seen in fig. \ref{fig:6}.

\subsection{ConvBlock} 
Our architecture is made of 8 ConvBlocks, each consisting of depthwise convolution (i.e. “mixing” spatial information) as in Multi-Layer Perceptrons (MLP-Mixer)\cite{Tolstikhin2021}, and Spatial Dropout\cite{Lee2020} for  strongly correlated pixels, followed by pointwise convolution (i.e. “mixing” the per-location features).
Each of the convolutions is followed by Gaussian error linear units (GELU)\cite{Hendrycks2016} activation and BatchNorm.
We found that for the task of keypoint estimation, architectures with a fewer number of layers result in better performance. We also added residual connections between Conv layers. We use a dropout with a rate of 0.2 to prevent overfitting.
The structure of our ConvBlock can be seen in the bottom panel of fig. \ref{fig:6}.

\subsection{Hybrid Prediction and a Multi-Task Loss Function} 

\acfp{FCN} are good at transforming one image to produce another related image, or a set of images while preserving spatial information.
Therefore, for our keypoint task, instead of using \ac{FCN} to directly predict a numerical value
of each keypoint coordinate as an output (i.e. regressing images to coordinate values), we modified \ac{FCN} to predict a stack of
output heatmaps (i.e. confidence maps), one for each keypoint.
The position of each keypoint is indicated by a single, two-dimensional, symmetric Gaussian in each heatmap in the output, and the scalar value of the peak reflects the prediction's confidence score.

Moreover, our network not only predicts heatmaps but also predicts scalar values for coordinates of each keypoint.
Therefore,  during the training process, we have a multi-task loss function, which consists of two losses, i.e. Jensen–Shannon divergence for heatmaps and Euclidean distance for coordinates (see figure \ref{fig:10}).
The first loss measures the distances between the predicted heatmaps and the ground-truth heatmaps, while the second loss measures the distances between the predicted coordinates and the ground-truth coordinates.
Then, we take the average of the two losses as the optimisation loss.

\section{Materials and methods} 

We ran three main experiments to test and optimize our proposed model.
First, we trained our network (MFLD-net) on only $40\%$ of our dataset. Next, we tested its predictive performance on the dataset test subset described below.
Finally, we compared our MFLD-net to five models from \cite{Ronneberger2015, He2015ResNet, Zhang2018b, MobileNetV2, Iandola2016}.
We assessed both the inference speed and prediction accuracy of each model as well as their training time and generalisability.
When comparing these models we incorporated the number of model parameters, the model size on the hard disk, and the model image throughput per second.
We applied the same configuration for each of the six investigated models in order to hold the training routine the same for all models.
The models are also trained using the same data augmentations, without affecting their performance.
The following sections are describing in detail the materials and methods used in the work.

\begin{figure}[htbp]
\centering
\includegraphics[width=0.45\textwidth]{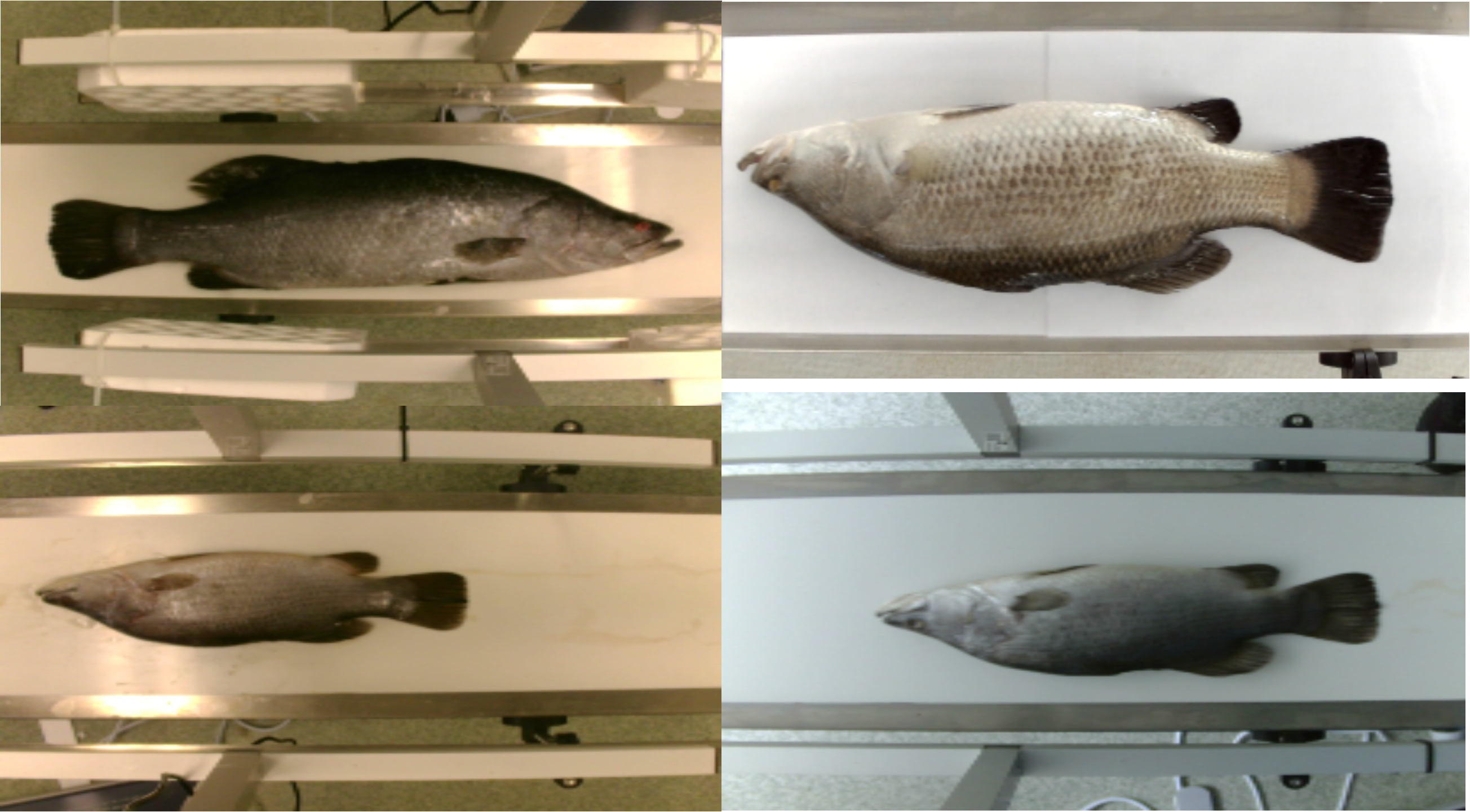}
\caption{Sample  images from the four data collection sessions, which are all used in our experiments.}
\label{fig:8}
\end{figure}

\begin{figure*}[htbp]
\centering
\includegraphics[width=0.98\textwidth]{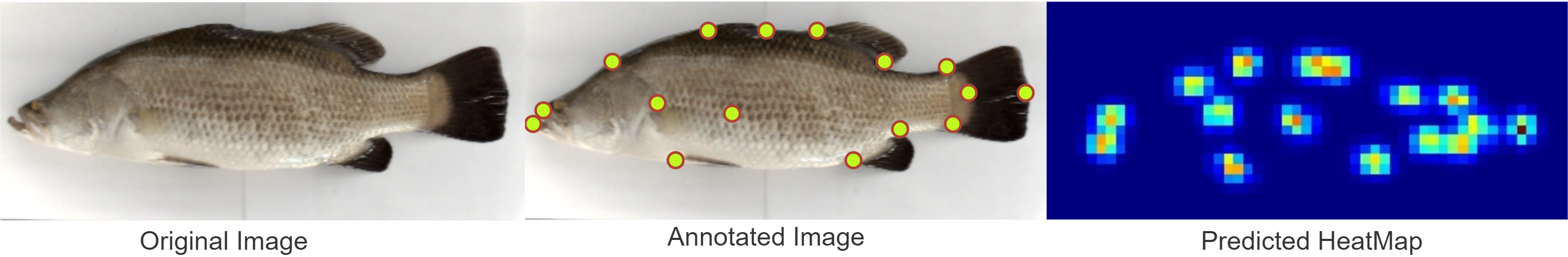}
\caption{Point annotations in a sample fish image ($\mathbf{X}$) (left). The points in the training ($\mathbf{Y}$) are inflated and highlighted for visibility, but only the centre pixel and its class label are collected and used (middle). The predicted Heatmap of the model  (right).}
\label{fig:3}
\end{figure*}

\subsection{Datasets} \label{secdata}
 
We performed experiments using a dataset of Barramundi (\textit{Lates calcarifer}), also known as Asian seabass. These fish were photographed in a laboratory setting.
The dataset was collected in four data collection sessions using the same experimental data collection setup but under two different environmental, i.e. lighting, conditions.
Figure \ref{fig:8} demonstrates a sample image from each of the four data collection trials. In total, 2500 images were collected, each of which was photographed on a conveyor belt with normal ambient lighting.
The images were recorded from above using a High-performance CMOS industrial camera (see figure \ref{fig:8}).
All barramundi were provided by the aquaculture team from James Cook University, Townsville, Australia.

To demonstrate the robustness of our network, we trained and validated our network on only $40\%$ of the dataset. This training subset was further split into randomly selected training and validation sets, with $70\%$ training examples and $30\%$ validation examples. The other $60\%$ of the collected dataset was used only for testing the model and comparing its performance to five other state-of-the-art baseline models \cite{Ronneberger2015, He2015ResNet, Zhang2018b, MobileNetV2, Iandola2016}.. 
The images were manually annotated for 16 kepoints as shown in figure \ref{fig:3}-middle.
For each fish, ground truth keypoints have the form $[(x_1,y_1),...,(x_k,y_k)]$, where $(x_i,y_i)$ represents the ith keypoint location. Each ground truth object also has a scale $s$ which we define as the square root of the object segment area. For each fish, our developed keypoint detector model outputs keypoint locations (see figure \ref{fig:3}-right). Predicted keypoints for each fish have the same form as the ground truth, i.e. $[x1,y1,...,xk,yk]$.

\subsection{Data Augmentation} \label{secaug}

To improve the training of our network and examine its robustness to rotation, translation, scale, and noise, we apply spatial and pixel level augmentation to our training data for all models using Albumentations library\cite{Buslaev2020}. In particular, we apply the following image transformations:
\begin{enumerate}
    \item Randomly flip an image horizontally with a probability of $0.5$.
    \item Randomly flip an image vertically with a probability of $0.5$.
    \item Randomly shift and scale an image with shift limit of $0.0625^{\circ}$, scale limit of $0.20^{\circ}$ with a probability of $0.5$.
    \item Randomly rotate an image with a rotation limit of $20^{\circ}$ with a probability of $0.5$. 
    \item Randomly blur an image with blur limit of $1$ with a probability of $0.3$.
    \item Randomly RGB-Shift an image with R-shift limit of $25$, G-shift limit of $25$, B-shift limit of $25$ with a probability of $0.3$. These augmentations help to further ensure robustness to shifts in lighting.
\end{enumerate}

We did not apply any of the image transformation operations to our validation or test sets.

\subsection{Performance Metrics} \label{secmetrc}

The following metrics were used to optimise and evaluate the model and to compare the quality of the predicted keypoint locations:

\paragraph{Euclidean distance}
measures the distance of the keypoints based on their coordinates (i.e  the line segment between the two points), and does not depend on how the ground truth has been determined \cite{Wang2005}. The best value of 0 indicates that the predicted keypoint is exactly at the same coordinate of the ground truth keypoint. 
 
We calculate the sum of the squared Euclidean distance of the difference between two feature vectors, i.e., the predicted feature vector and the ground truth feature vector. 
This represents the total difference between the two feature vectors. The Euclidean distance is

\begin{equation}
d(g,p)= \sqrt{\sum_{i=1}^n  ( v_{i}^{g} - v_{i}^{p} )^{2}},
\label{eq:Euclidean}
\end{equation}
where $g$ and $p$ are two sets of points in Euclidean n-space for ground truth and prediction, respectively. $ v_{i}^{g} , v_{i}^{p}$ are	Euclidean vectors, starting from the origin of the space (initial point) for the ground truth and prediction, respectively. $n$ is the number of keypoints.

\paragraph{Jensen-Shannon divergence}
is a distance measure between two distributions, such as the difference between the predicted and ground truth point distributions \cite{Nielsen2020}. It can therefore be used to quantify the accuracy of the predicted keypoints. The lower this value is, the better the model performs.
 
This distance is calculated based on the Kullback-Leibler divergence (KLD)\cite{Contreras-Reyes2012}, where the inputs for the summation are probability distribution pairs. The $KLD$ for two probability distributions, $P$ and $Q$ and when there are $n$ pairs of predicted $p$, and ground truth $g$, can be expressed as:
\begin{equation}
KLD(P||Q) =  \sum_{i=1}^n p_i(x)log \left(\frac{p_i(x)}{q_i(x)}\right),
\label{eq:kld}
\end{equation}
to measure the difference between two probability distributions over the same
variable $x$ and indicate the dissimilarity between the distributions. The best value is $0$. Utilising $KLD$, $JSD$ can be expressed as follows: 
\begin{equation}
JSD_M(P||Q) = \sqrt{\frac{KLD(p \parallel m) + KLD(q \parallel m)}{2}},
\label{eq:Jensen}
\end{equation}
where $m$ is the point-wise mean of $p$ and $q$.

This is a measure of the difference between two probability distributions $P$ and $Q$. As can be seen from the formula, the best value of 0 indicates no difference between the distributions.

\paragraph{Object Keypoint Similarity (OKS)}
OKS keypoints estimation serves the same purpose as Intersection over Union ($IoU$) as in object detection.
It is determined by dividing the distance between expected and ground truth points by the object's scale\cite{Lin2014b}.
This gives the similarity between the keypoints (or corners) of the two detected boxes. The result is between 0 and 1, where 0 means no similarity between the keypoints, while perfect predictions will have OKS=1. The equation is as follows: 

\begin{equation}
OKS = \frac{\sum_{i} \exp \left(-d_{i}^{2} / 2 s^{2} k_{i}^{2}\right) \delta\left(v_{i}>0\right)}{\sum_{i} \delta\left(v_{i}>0\right)},
\label{eq:oks}
\end{equation}
where $d_{i}$ is the Euclidean distance between the detected keypoint and the corresponding ground truth, 
$v_{i}$ is the visibility flag of the ground truth, $s$ is the object scale, while $k_{i}s$ represents a per-keypoint constant that controls falloff.

To compute OKS, we pass the $d_{i}$ through an unnormalized Gaussian with standard deviation $k_{i}s$. 
For each keypoint, this yields a keypoint similarity that ranges between 0 and 1. These similarities are averaged over all labelled keypoints. 
Given the OKS, we can compute Average Precision ($AP$) and Average Recall ($AR$) just as the $IoU$ allows us to compute these metrics for box/segment detection.

Both equations \ref{eq:Euclidean} and \ref{eq:Jensen} have been used for model training and optimisation, and also used to compare different models' performance as in table \ref{tab:Params}.
Equation \ref{eq:oks} was used as a final evaluation metric for all the models used in this study.

\begin{figure}[htbp]
\centering
\includegraphics[width=0.48\textwidth]{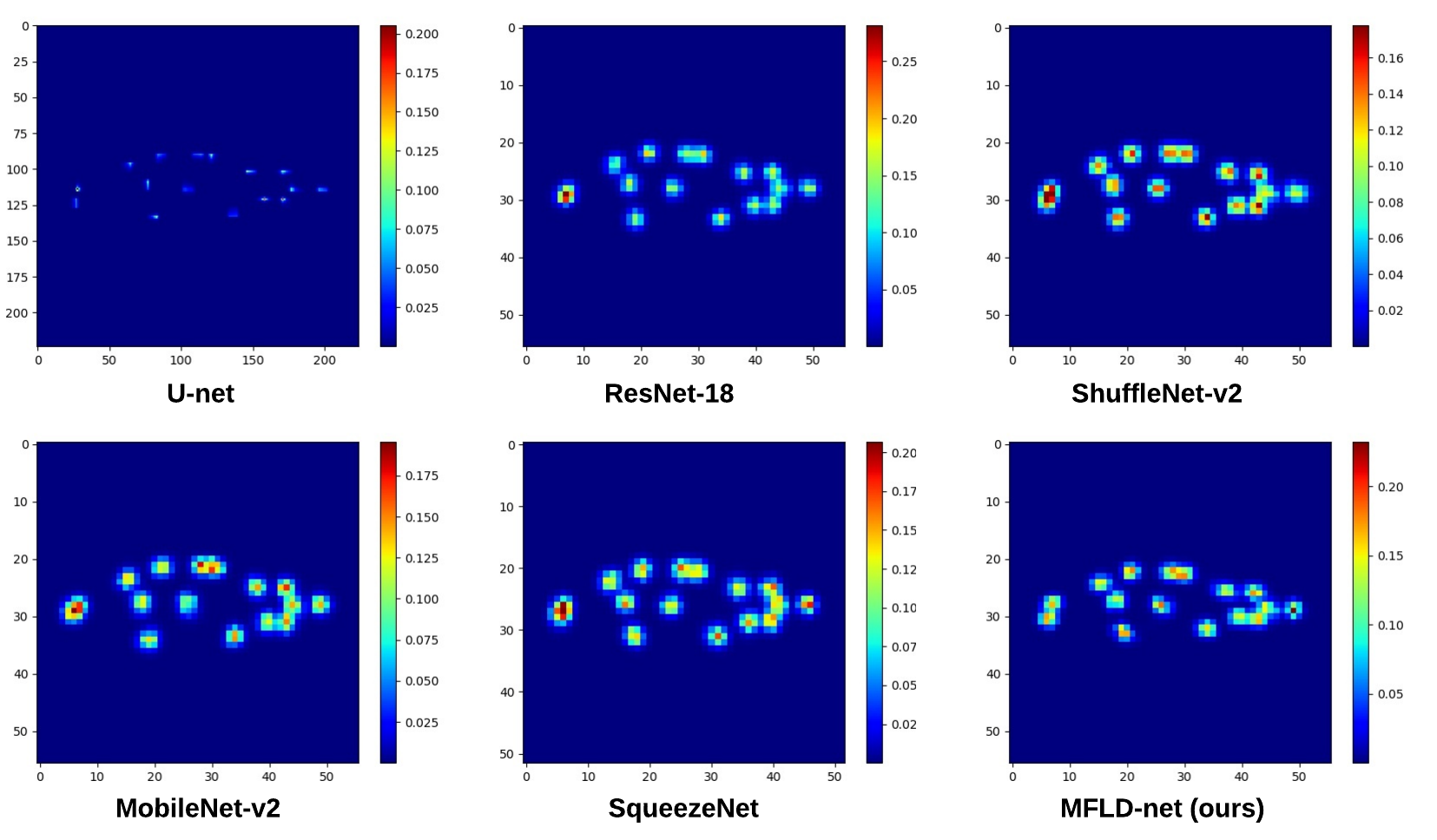}
\caption{Sample output heatmap from each of the 6 networks used in this work.}
\label{fig:2}
\end{figure}

\subsection{Model training} \label{sectrain}

We trained six different models on the training subset. The models used for training are
U-net\cite{Ronneberger2015}, ResNet-18\cite{He2015ResNet}, ShuffleNet-v2\cite{Zhang2018b}, MobileNet-v2\cite{MobileNetV2}, SqueezeNet\cite{Iandola2016} and our proposed lightweight network MFLD-net.
For each experiment, we set our model hyperparameters to the same configuration for all models.  All the models were trained with $224\times224$ resolution input and  $56\times56$ heatmap resolution output except U-net\cite{Ronneberger2015} with $224\times224$ resolution for both input and output (see figure \ref{fig:2}). Each model has two outputs (heatmap and coordinates), where two losses were applied as shown in figure \ref{fig:10}.

We found that for this problem set, a learning rate of $1 \times 10^{-3}$ works the best. It took around 50 epochs for all models to train on this problem and the learning rate was decayed by $\gamma = 0.1$  every 30 epochs. Our networks were trained on a  Linux host with a single NVidia GeForce RTX 2080 Ti GPU using Pytorch framework\cite{Paszke2019}. The batch size we used was $64$. We used Adam optimiser\cite{Kingma2014Adam:Optimization} with $\beta_1 = 0.9$, $\beta_2 = 0.999$, and $\epsilon = 1.0 \times 10^{-08}$.
We applied the same hyperparameter configuration for all six models.
The optimum model configuration will depend on the application, hence, these results are not intended to represent a complete search of model configurations.

Because we only used the training subset ($n=1000$ images) for training and validation, during optimization, we heavily augmented our training set, challenging the model to learn a much broader data distribution than that in the training set.
We applied several image transformations for data augmentation as specified in section \ref{secaug}.

\begin{figure}[htbp]
\centering
\includegraphics[width=0.40\textwidth]{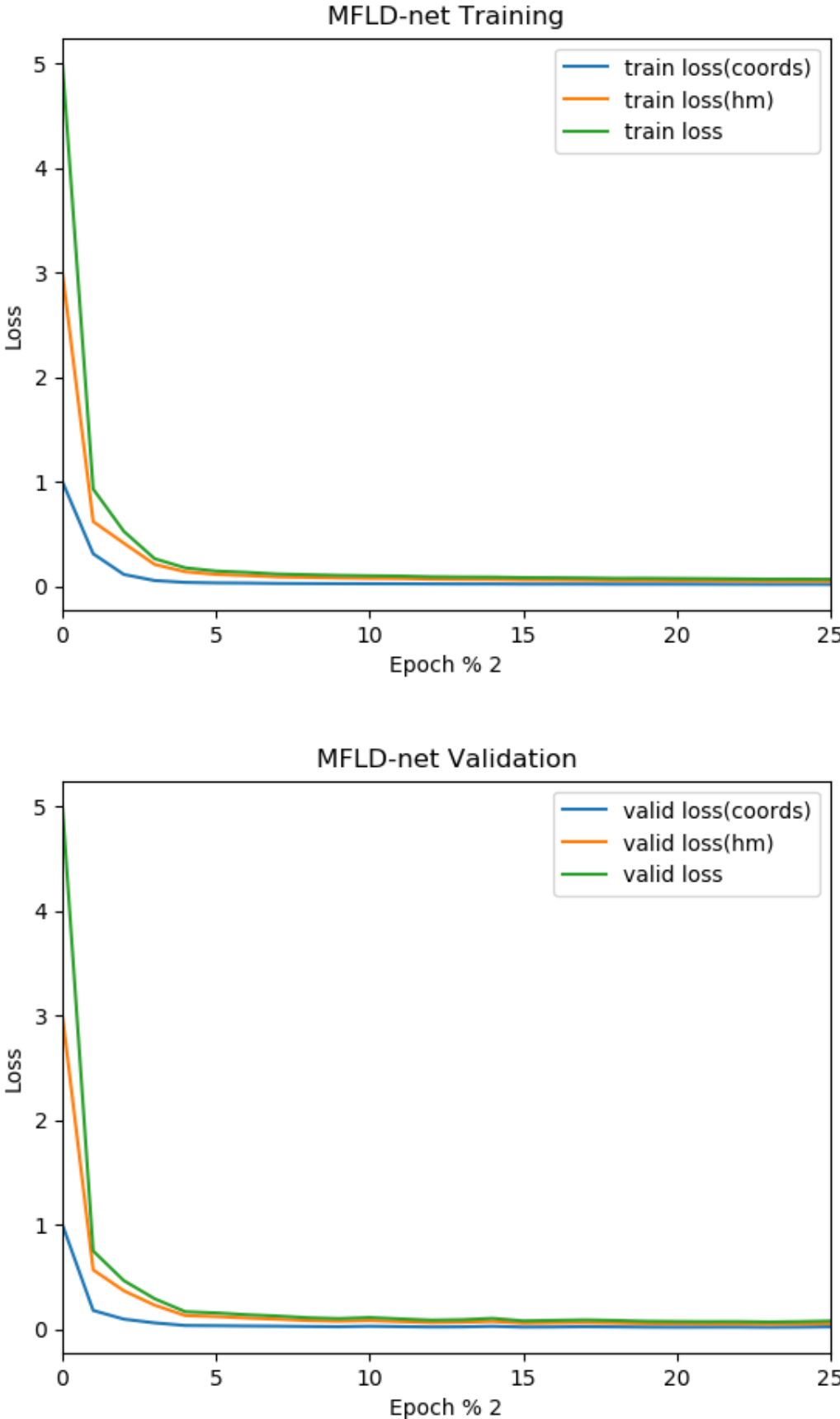}
\caption{The two different losses, i.e. coordinate and heatmap prediction losses are shown along with the total loss for both training and validation.}
\label{fig:1}
\end{figure}

We regarded the model to be converged when the validation loss stopped improving after 50 epochs.
Only for the best performing version of the models, we calculated validation error as the Euclidean distance between predicted and ground-truth picture coordinates and Jensen-Shannon divergences between heatmaps and centres of the target Gaussians, which we assessed at the end of each epoch during optimization.
Figure \ref{fig:1} shows the training and validation losses for our proposed network.

\subsection{Model evaluation} 

\acf{DL} models are typically evaluated for their predictive performance (i.e. ability to generalize to new data), using a sub-sample of annotated data (test set) that is not used for training or validation.
A test set is typically used to avoid overfitting the model hyperparameters to the validation set, which can result in biased performance measurements.
Therefore, we used only $40\%$ of our dataset for training  and left the other $60\%$ of its images for testing the model's predictive performance using metrics described in section \ref{secmetrc}.

\begin{figure*}[htbp]
\centering
\includegraphics[width=0.98\textwidth]{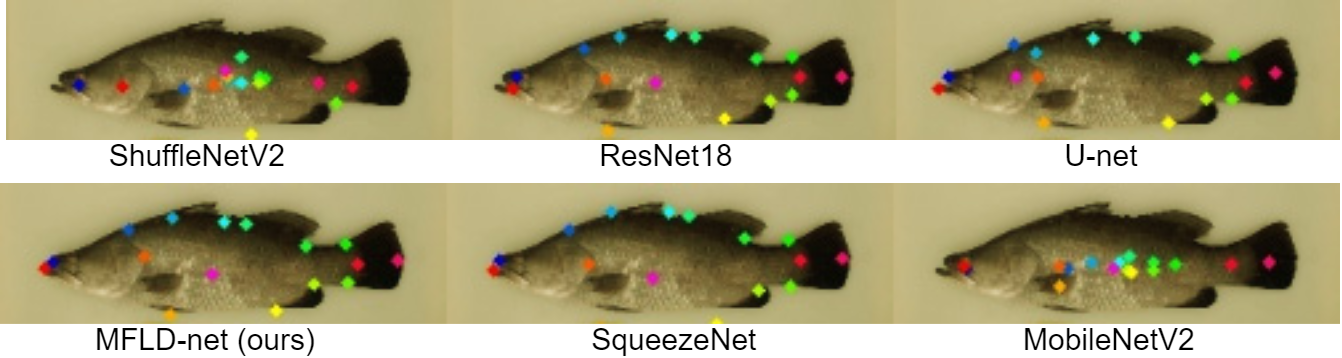}
\caption{Example keypoints estimation predicted by the proposed network and a state of the art CNNs.}
\label{fig:7}
\end{figure*}

\section{Results} \label{secresult}

To fully evaluate our model and compare it with other methods, we ran experiments to optimize our approach and compared it to the five aforementioned models 
in terms of image throughput (speed), accuracy, inference time, and generalization ability. 
We benchmarked these models using the test subset (see section \ref{secdata} for details).

We applied the same training configuration for all of the six models, 
meaning that the models are all trained using the same dataset and data augmentations as explained in section \ref{sectrain}.

\subsection{Performance Comparison}
Table \ref{tab:Params} shows comparative results based on the number of parameters of a model, the model size on the hard disk, and the model throughput in image per second. In addition, the coordinates loss (Equ. \ref{eq:Euclidean}), heatmap loss (Equ. \ref{eq:Jensen}), and the average of both losses are shown.
All the tests were conducted on a desktop computer with a single NVidia GeForce RTX 2080 Ti GPU.
 
\begin{table}[htbp]
\centering
\caption{Performance Comparison to other models.}
\resizebox{\linewidth}{!}{%
\begin{tabular}{l|ccc|ccc}  
\toprule
Network                      & \shortstack{\# Params \\ (x$10^6$)}   & \shortstack{Size \\ (MB)}  & \shortstack{Throughput \\ (img/sec)} & Coords & HeatMap & Avg. \\ 
\midrule
U-net\cite{Ronneberger2015}  & 31.04 &   124.3     &    201     &  0.024 &  0.355   & 0.190    \\ 
ResNet-18\cite{He2015ResNet} & 12.85 &   51.5    &     404  &  0.028 &   0.090   &  0.059   \\  
ShuffleNet-v2\cite{Zhang2018b}& 3.06 &   12.5     &    170     &  0.047 &   0.153  &  0.100   \\  
MobileNet-v2\cite{MobileNetV2}& 4.10 &   16.7    &    205    &  0.041 &   0.137   &  0.089   \\  
SqueezeNet\cite{Iandola2016} & 2.33 &   9.4    &    551    &  0.027 &   0.078   &  0.052   \\  
\midrule
\textbf{MFLD-net (ours)}   & \textbf{0.65} &   \textbf{2.7}     &    480      &  0.039 &   0.120   &  0.080   \\  
\bottomrule
\end{tabular}
}
\label{tab:Params}
\end{table}

Overall, the results summarized in table \ref{tab:Params} show that our network (MFLD-net) outperforms other networks, achieving the lowest number of parameters (47x fewer parameters than U-net\cite{Ronneberger2015}), the smallest size on the hard disk, and the second-highest throughput after SqueezeNet\cite{Iandola2016}.
Also, our model has a lower average loss than U-net\cite{Ronneberger2015}, ShuffleNet-v2\cite{Zhang2018b}, and MobileNet-v2\cite{MobileNetV2}.
The small number of parameters as well as the very compact size of our model while having a high throughput makes it an appealing solution for many problems such as real-time mobile fish video processing and portable autonomous systems \cite{saleh2022a}.

\subsection{Precision and Recall}

We also show in table \ref{tab:results} that our MFLD-net model achieves good generalisation with few training examples and without the use of transfer learning when combined with strong data augmentation.
To further examine the efficacy of our model generalisation, we compared the performance of our model with randomly initialized weights, against the other models initialized with weights pre-trained on the ImageNet dataset\cite{Li2009}.

Object Keypoint Similarity (OKS)\cite{Lin2014b} was used as a performance metric (see section \ref{secmetrc} for more details).
As explained, the following 6 metrics are usually used for characterising the performance of a keypoint detector model. We, therefore, used them. 
\begin{itemize}
    \item Average Precision ($AP$):
        \begin{itemize}
            \item $AP$ ~~ ( at $OKS=.50:.05:.95$ (primary metric))
            \item $AP^{.50}$ ( at $OKS=.50$ )
            \item $AP^{.75}$ ( at $OKS=.75$ )
        \end{itemize}
    \item Average Recall ($AR$):
        \begin{itemize}
            \item $AR$ ~~ ( at $OKS=.50:.05:.95$)
            \item $AR^{.50}$ ( at $OKS=.50$)
            \item $AR^{.75}$  ( at $OKS=.75$)
        \end{itemize}
\end{itemize}

\begin{table}[htbp]
\centering
\caption{Performance comparison using the $OKS$ metric on the \textbf{test} datasets.}
\resizebox{\linewidth}{!}{%
\begin{tabular}{l|ccc|ccc}
\toprule
Network                      & $AP$ & $AP^{.50}$ & $AP^{.75}$ & $AR$ & $AR^{.50}$ & $AR^{.75}$   \\ 
\midrule
U-net\cite{Ronneberger2015}  & 0.981 &   0.990   &  0.990      &  0.994 &  0.999    & 0.999    \\ 
ResNet-18\cite{He2015ResNet} & 0.984 &   0.990   &   0.990     &  0.996 &   0.999   &  0.999   \\  
ShuffleNet-v2\cite{Zhang2018b}& 0.957 &   0.990   &   0.990     &  0.979 &   0.999   &  0.999   \\  
MobileNet-v2\cite{MobileNetV2}& 0.963 &   0.990   &   0.989     &  0.979 &   0.999   &  0.996   \\  
SqueezeNet\cite{Iandola2016} & 0.971 &   0.990   &   0.990     &  0.988 &   0.999   &  0.999   \\  
\midrule
\textbf{MFLD-net (ours)}      & 0.967 &   0.990   &   0.990     &  0.983 &   0.999   &  0.999   \\  
\bottomrule
\end{tabular}
}
\label{tab:results}
\end{table}

Overall, the results summarized in table \ref{tab:results} shows that our network (MFLD-net) outperforms both ShuffleNet-v2\cite{Zhang2018b}, and MobileNet-v2\cite{MobileNetV2}, achieving $AP=0.967$, while being competitive with SqueezeNet\cite{Iandola2016}, 
despite having substantially fewer parameters.
More importantly, we achieve this high accuracy with a small network (0.65M parameters only), but without the use of transfer learning. This shows the effectiveness and generalisability of our MFLD-net model.

\subsection{Qualitative Results}

To further confirm our model generalization on  the unseen images, we perform a qualitative experiment on the test subset, with sample results shown in figure \ref{fig:7}. This figure clearly shows that our network performs better than the previous methods. The other methods have the problem of misclassifying the pixels with a similar intensity of one colour as the other colour, whereas our method shows a strong ability to differentiate pixels with similar intensity. We can also clearly see that the proposed method can work on images with different lighting conditions.

\section{Discussion} 

The main goal of this study was to create a convolutional neural network that can achieve competitive or better results in low data regimes and be more feasible for usage in embedded and mobile devices to estimate keypoints or landmarks on objects. We applied our model to an important research and industry application, i.e. fish morphology.

\begin{figure}[htbp]
\centering
\includegraphics[width=0.48\textwidth]{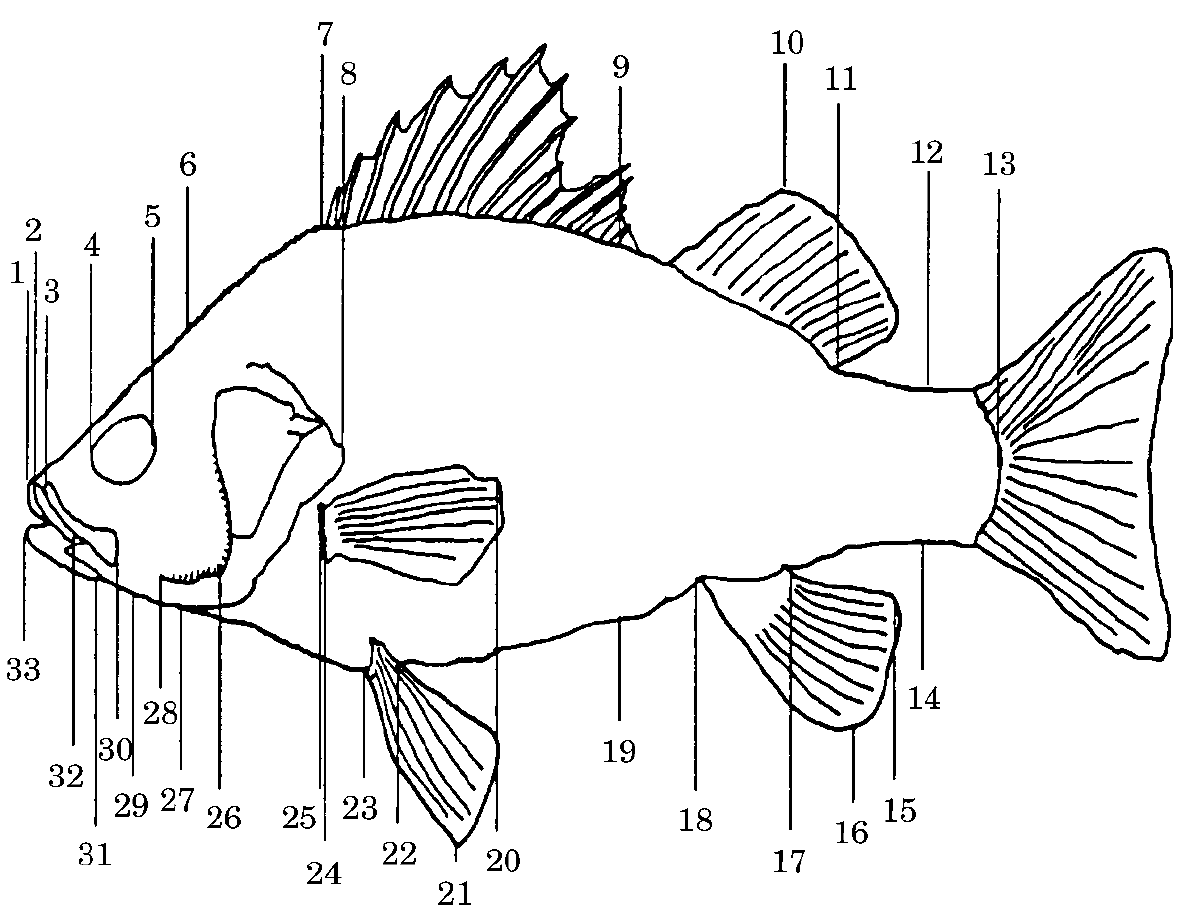}
\caption{Position of most of the landmark points used to describe shape variation in \textit{M. novemaculeata} from seven geographically distinct rivers. See \cite{Jerry1998} for an explanation of variables measured. The figure from \cite{Jerry1998}.}
\label{fig:11}
\end{figure}

A common and important practice in aquaculture is determining the morphology of fish (see figure \ref{fig:11}). Fish morphology determination is required for both selecting and evaluating novel fish strains for cultivation. The most widely used method to characterise fish is by observation of their overall appearance. An experienced observer can determine a fish's size, weight, possibly sex, and even its condition. 
The traditional observation method to evaluate fish morphology includes weighing fish,  measuring lengths with a ruler or calipers and or some other aspect of the fish, and then recording these observations on a form (usually a piece of paper). This observation process is slow, labour-intensive, and highly prone to human error. 

A possible solution could automate the fish observation process if an accurate mobile system is developed that can be deployed in the field and in fish farms. 
This fish morphometric tool could quickly measure various fish features and morphological traits from fish images captured online or offline using a camera. The tool also collects the morphological data, and then uses it for analysis and producing a final report. 
Such a tool is very useful to aquaculture and fish farms and could provide a new way to select, evaluate, and analyze fish and other aquaculture animal products. 

In this paper, we developed a novel deep learning algorithm for accurate fish morphometric measurements from fish images. To efficiently measure various fish morphological traits, we developed 
a fish-specific landmark detection model that could accurately localise keypoints (landmarks) on the fish body (for example see figure \ref{fig:11}. These landmarks can be then used to rapidly measure various fish traits including their weight, length, head shape, and body shape. In addition, the fish landmarks can be used to describe shape variation, deformation and differential development in various fish species \cite{Jerry1998,Jerry2022bara}.

We build our land-mark detection model upon the most widely-used deep learning variant, i.e. \ac{CNN}.  
A number of factors can significantly influence \acp{CNN} performance. These factors include the size of the network (including the number of layers, number of kernels, and their width), the number of input features, and the size of the training set. In addition, the use of the convolution layers affects the size and complexity of the network but can help to decrease the error rate and improve prediction accuracy. However, there is no clear mechanism to arrive at the optimal convolutional architecture for a specific task. The architecture selection involves choosing important hyperparameters such as the network structures and the training time. 

Through experimentation and using our experience in developing deep learning algorithms, we designed a lightweight \ac{CNN} with a short training time and high generalisability to make it suitable for fast deployment and real-time mobile applications in fish farms. 
Our experiments showed that MFLD-net best performances can be achieved by (i) increasing the size of the kernel to 9, (ii) including more input dimensions by Patch embeddings, and (iii) reducing the number of convolution layers to 8. The reduction in the number of convolution layers resulted in a model with fewer parameters that achieved better generalisation capabilities compared to the state-of-the-art models.

To train and evaluate our model performance, we collected a dataset containing $2500$ harvested or sedated fish images. These images were manually annotated for important landmarks on the fish body. We used a combination of data augmentation techniques to improve the network’s performance in a low data regime. In our experiments, the input images were scaled to a size of $224\times224$, and the output was the position of each fish landmark. These landmarks (keypoints) were indicated by a single, two-dimensional, symmetric Gaussian heatmap, where a scalar peak value reflects the prediction's confidence score.
The quantitative and qualitative experimental results showed that our proposed model while being significantly lighter, can outperform some and be competitive with other state-of-the-art models. We also showed that our model has a high generalisation capability and does not need transfer learning even when using a small training dataset.

The main limitation of our study is that all the samples for training and testing are taken from a similar source, even though, they were slightly different, due to being collected in different conditions and by different operators. Another limitation is the use of a single fish species in our dataset.
Since there are a variety of different species and sizes of fish in the aquaculture industry, there is a need to test the model for more than one species.
However, our aim in this study was to build a proof of concept, which can be extended in future works to other species. 
Our presented results indicate that our developed MFLD-net model trained using images from a single species could be generalised to detect fish of different species and in different environments. This could be the subject of future research.
In addition, in future work, the model can be trained with images of other objects, or images captured from different fish species. It is worth noting that, collecting new fish images and annotating them is a time-consuming and expensive exercise. This was the case, even in our data collection trials, where fish images were collected when the fish passed on a conveyor belt and under a camera capturing videos.

In addition, developing new low-cost, low-power, and high-speed mobile devices has been an evolving research area in many applications such as agriculture\cite{Lammie2019}, and marine science\cite{Jahanbakht2021,Jahanbakht2022a}. These devices need a lightweight and fast networks, such as
the proposed model in this work. Therefore, an interesting future research project is to develop a low-cost mobile device to perform fish morphology estimation using the proposed network.

\section{Conclusion}
This paper presented a lightweight transformer-based CNN for detecting object landmarks (keypoints). The developed network was applied to a fish landmark detection task with the aim to have a model that can be trained with a small number of training samples and can generalise to new unseen images.  
We trained our model on $40\%$ of our fish image dataset, which was collected with different light conditions. We then evaluated it on the other $60\%$ of the dataset. We compared our method to five state-of-the-art baseline models and showed that it is better or on par with these models while being significantly lighter than all of them.
This work can pave the way for future research in developing lightweight landmark detection CNNs toward intelligent low-power processors to be used on mobile devices such as drones and robots.

\section*{Acknowledgement}
This research is supported by an Australian Research Training Program (RTP) Scholarship and Food Agility HDR Top-Up Scholarship. D. Jerry and M. Rahimi Azghadi acknowledge the Australian Research Council through their Industrial Transformation Research Hub program.

	
\ifCLASSOPTIONcaptionsoff
\newpage
\fi

\bibliographystyle{IEEEtran}
\bibliography{references}

\end{document}